\documentclass{article}

\usepackage{spconf,amsmath,graphicx}
\usepackage{booktabs,multirow}
\usepackage{amssymb}
\usepackage{hyperref}
\usepackage{cleveref}
\usepackage{xcolor}

\usepackage{graphicx}
\usepackage{array}
\usepackage{float}
\usepackage{cite}
\usepackage{colortbl} 
\usepackage{enumerate}
\usepackage{bbm}

\let\high\relax
\let\Tstrut\relax

\title{DAP: Domain-aware Prompt Learning for Vision-and-Language Navigation}
\name{Ting Liu$^{1}$
Yue Hu$^{1}$
Wansen Wu$^1$
Youkai Wang$^1$
Kai Xu$^{1}$
Quanjun Yin$^{1\dag}$
\thanks{$^\dag$Corresponding author.}}
\address{$^1$College of Systems Engineering, National University of Defense Technology, Changsha, China}

\begin{document}

\maketitle
\begin{abstract}
Following language instructions to navigate in unseen environments is a challenging task for autonomous embodied agents. With strong representation capabilities, pretrained vision-and-language models are widely used in VLN. However, most of them are trained on web-crawled general-purpose datasets, which incurs a considerable domain gap when used for VLN tasks. To address the problem, we propose a novel and model-agnostic \textbf{D}omain-\textbf{A}ware \textbf{P}rompt learning (\textit{DAP}) framework. For equipping the pretrained models with specific \textit{object-level} and \textit{scene-level} cross-modal alignment in VLN tasks, DAP applies a low-cost prompt tuning paradigm to learn soft visual prompts for extracting in-domain image semantics. Specifically, we first generate a set of in-domain image-text pairs with the help of the CLIP model. Then we introduce soft visual prompts in the input space of the visual encoder in a pretrained model. DAP injects in-domain visual knowledge into the visual encoder of the pretrained model in an efficient way. Experimental results on both R2R and REVERIE show the superiority of DAP compared to existing state-of-the-art methods.
\end{abstract}
\section{Introduction}
\label{sec:intro}

Recently, embodied artificial intelligence \cite{DBLP:conf/icassp/LiSWC22} has attracted extensive attention in various kinds of robots and virtual agents. An embodied agent learns by observing, moving, talking, and interacting in an active learning scenario, similar to what humans do in the real world. There has been a large family of related tasks proposed in this field. In particular, vision-and-language navigation (VLN)~\cite{qi2020reverie,majumdar2020improving} is a task where the agent is required to navigate following a natural language instruction in a photo-realistic simulated environment.

Like some CV tasks built on pretrained visual models, such as 3D Semantic Scene Completion~\cite{yao2023ndc,yao2023depthssc}. A large number of VLN methods~\cite{majumdar2020improving,DBLP:conf/cvpr/HaoLLCG20} are also built on pretrained vision-and-language models, which are usually learned from web-crawled datasets. In general, there are two major processes in these approaches, namely pretraining and/or fine-tuning a representation model, and inferring the navigation actions based on the pretrained multi-modal representation model. The effectiveness of such methods is largely affected by the representation ability of pretrained models\cite{liu2023vgdiffzero}. In this regard, a successful and efficient VLN agent entails the understanding of the in-domain scene semantics. However, most pretrained models are trained on web-crawled image-text pairs, resulting in a considerable domain gap between pretraining datasets and VLN datasets. It is hard for them to efficiently align the semantics in specific visual scenes and textual instructions. This kind of disability inhibit the agent's ability to reason about in-domain visual scenes accurately. 

Prompt learning is a recently popular learning paradigm that is usually used with pretrained large-scale models. It processes the input information based on a prompt template, which embeds the expected output into a fill-in-the-blank format. In this way, downstream tasks can be reconstructed into a form that can better utilize pretrained models. Prompt learning has been extensively studied with significant success in Natural Language Processing (NLP)~\cite{brown2020language} and Computer Vision (CV)~\cite{radford2021learning,DBLP:journals/corr/abs-2109-11797}, and has also been introduced into VLN tasks~\cite{DBLP:conf/cvpr/Lin0CLLL22,DBLP:conf/acl/LiangZLXL22} in some latest works. For example, ProbES~\cite{DBLP:conf/acl/LiangZLXL22} generates instructions corresponding to sampled paths by prompt templates, which builds a large-scale VLN dataset for pretraining. ADAPT~\cite{DBLP:conf/cvpr/Lin0CLLL22} generates an action prompt base via the CLIP~\cite{radford2021learning} model to enable the explicit learning of action-level modality alignment and achieve some performance improvement. However, they ignore the domain gap between pretraining datasets and VLN datasets. 

Motivated by these works, we endeavor to exploit prompt learning to bridge the domain gap between pretraining datasets and VLN datasets. To this end, we propose an original and model-agnostic framework termed Domain-aware Prompt Learning (\emph{DAP}). The core idea of DAP is to inject soft visual prompts, which are learned from in-domain scene, into a general pre-trained vision-and-language model. The learned prompts serve as domain-specific guidance that adapt the model to VLN tasks in an efficient and effective manner.

Specifically, to narrow the above domain gap, we first generate a set of in-domain image-text pairs as a supervised training dataset. Then we introduce soft visual prompts in the input space of the visual encoder in a pretrained model. The aim is to enhance its representation ability on specific \textit{object-level} and \textit{scene-level} cross-modal alignment, i.e., to enable it to identify the objects and scenes in VLN scenes, represented visually in the in-domain images, in linguistic forms. Only the visual prompts and an MLP head are learnable during training with the in-domain dataset, while the parameters

\begin{figure*}[ht]
\centering
\vspace{-2mm}
\includegraphics[width=\textwidth]{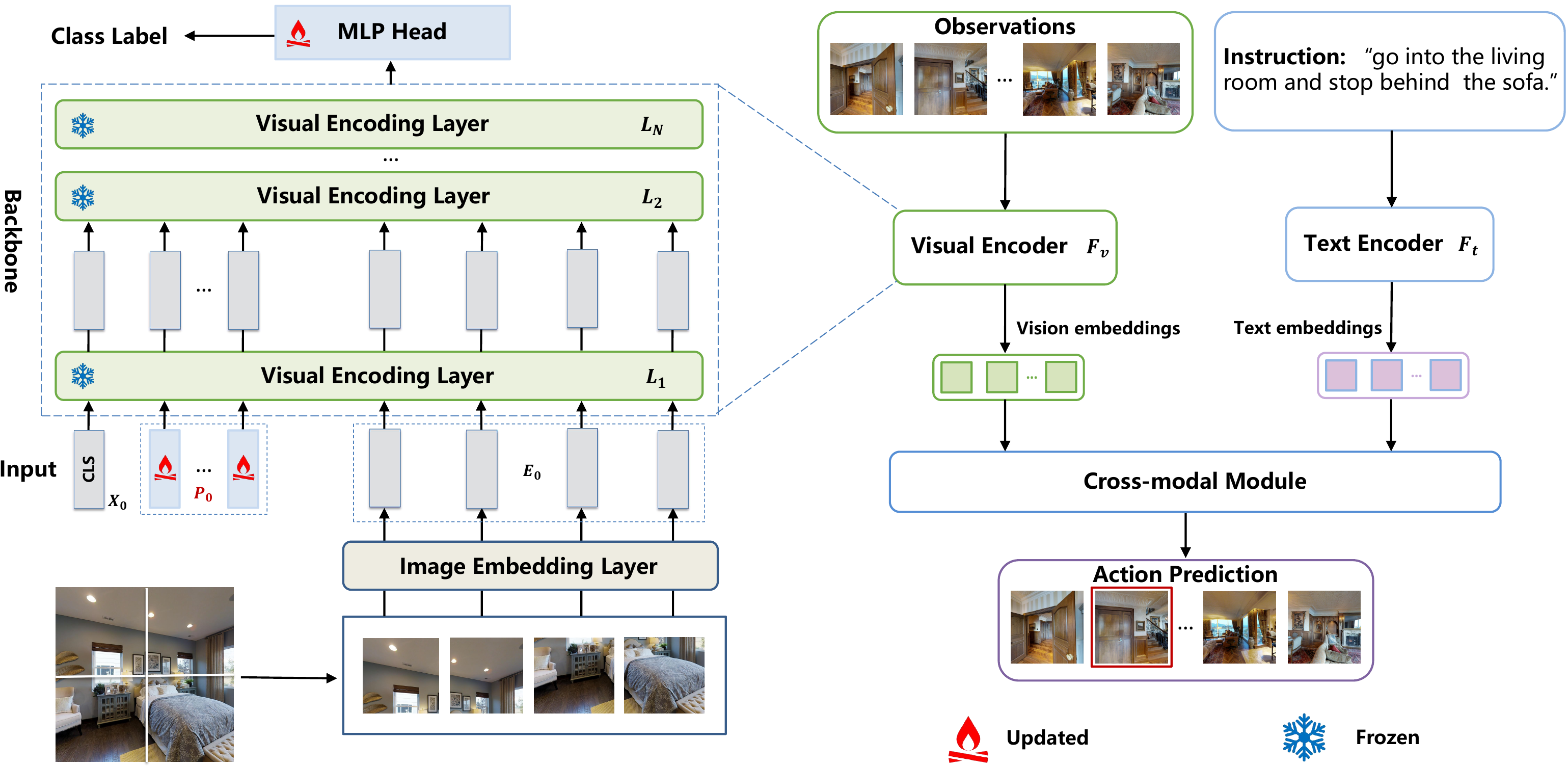}
\vspace{-3mm}
\caption{The overview of DAP pipeline for VLN. Soft visual prompts are inserted into the input space of the vision encoder, where only the parameters of soft visual prompts and the MLP head are updated during training. DAP focuses on learning soft visual prompts to enhance the adaptation of backbone models to VLN tasks.The visual-and-language model is used for action prediction after the visual module is injected with in-domain knowledge.}

\label{fig:fig2}
\end{figure*}

\noindent
of the pretrained model are kept frozen. With this fast and low-cost training paradigm, soft visual prompts learned from the in-domain dataset can adapt the pretrained models to VLN scenes very efficiently. It's worth noting that the prompt-based pipeline proposed in this paper is (i)
independent of specific large-scale pretrained vision-and-language models, and (ii) also orthogonal to models used for
navigation inference as long as they are designed based on
pretrained models.

In summary, the contributions of DAP are summarized as: (i) We present DAP to retrain the visual encoder of a vision-and-language representation model for VLN tasks that captures the in-domain scene semantics ; (ii) We introduce prompt learning as a fast and low-cost way to adapt pretrained models to VLN tasks; (iii) DAP shows promising performances and generalization ability with the help of prompt learning, and outperforms existing SOTA methods.

\section{Method}
\label{sec:Method}

\subsection{VLN Problem Setup}
\label{VLN Problem Setup}
The VLN agent is put in a photo-realistic environment, and it is assigned a random initial position and given a language instruction $I$. The VLN agent is required to find a path from the initial position to the target position following the instruction. At each time step $t$, the agent observes the environment and makes a current action decision that updates the agent state $s_{t}$ to a new state $s_{t+1}$. The state includes historical information and current spatial information consisting of a viewpoint and orientation. All viewpoints are on the connectivity graph $G = \left\langle\\V, E\right\rangle$ of the environment \cite{DBLP:journals/corr/abs-1807-06757}, where $V$ and $E$ represent navigable nodes and edges, respectively.

\vspace{-2mm}

\subsection{Prompt Engineering}
 To narrow the domain gap, We introduce learned visual prompts for adapting pretrained vision-and-language models to VLN tasks. we adopt the supervised learning method to learn soft visual prompts, taking in-domain images as the input and the text of the corresponding object as labels. We introduce a set of continuous embeddings, i.e., soft visual prompts, to serve as additional input. Soft visual prompts are automatically learned from an in-domain dataset by prompt tuning, while the parameters of pretrained models are kept frozen. These prompts assist the VLN agent to ground the object and scene descriptions in the instruction onto the visual perception. Each visual prompt is a learnable $d$-dimensional vector. As shown in Figure \ref{fig:fig2}, the form of learnable visual prompts encoded by the first visual encoding layer are continuous embeddings that can be represented as $\mathbf{P}_{0}$.

\vspace{-3mm}



\subsection{Text Generation with the CLIP Model}
The powerful cross-modal pre-trained model CLIP~\cite{radford2021learning} has a unique ability to generate text pseudo labels from VLN scene categories. This will guide us to design a supervised task for optimizing the learnable visual prompts. For the supervised signals, we first randomly sample a number of indoor images from the Matterport 3D dataset, then we apply the CLIP model to automatically generate pseudo text labels corresponding to sampled indoor images. The text encoder in the CLIP embeds the prompt template \textit{``A photo of a $\left\{object\right\}$"}, where the object represents the object class (e.g. chairs) or scene category (e.g. bedroom) corresponding to an indoor image, and the image encoder embeds the image to be predicted. Finally, the similarity between text embedding and image embedding is calculated, and we choose the text with the highest matching score for the image.


\subsection{Domain-aware with Soft Visual Prompts}
Soft visual prompts can be widely used in vision-and-language pretrained models to better understand in-domain image semantics. We apply the pretrained model PREVALENT \cite{hao2020towards} for demonstration. We inject indoor visual knowledge into the visual encoder of the pretrained PREVALENT model by prompt tuning. We introduce a set of soft visual prompts in the input space after the input images are initially processed by the embedding layer. Visual prompts are learnable during prompt tuning, while the PREVALENT backbone is kept frozen. Therefore, the pretrained model is quickly adapted to VLN tasks by the low-cost prompt tuning paradigm.

Soft visual prompts are inserted into the input space of the $N$-layer vision encoder in the PREVALENT. The output of each visual encoding layer is formulated as:

\begin{small}
    \begin{equation}
         \left[{\mathbf{X}}_{1},\mathbf{P}_{1},\mathbf{E}_{1}\right] = L_{1}\left(\mathbf{X}_{0},\mathbf{P}_{0},\mathbf{E}_{0}\right)
    \end{equation}
\end{small}

\vspace{-0.5cm} 
\begin{small}
    \begin{equation}
         \left[{\mathbf{X}}_{i},\mathbf{P}_{i},\mathbf{E}_{i}\right] = L_{i}\left(\mathbf{X}_{i-1},\mathbf{P}_{i-1},\mathbf{E}_{i-1}\right), \quad i=2,3,\ldots,N
    \end{equation}
\end{small}

\noindent
where $\mathbf{X}_{i}$, $\mathbf{P}_{i}$, and $\mathbf{E}_{i}$ denote the $\left[CLS\right]$, prompts and image features respectively encoded by the $i\mbox{-}th$ visual encoding layer. The output $\mathbf{X}_{N}$ is mapped by an MLP head to a predicted class probability distribution $y$. 

The PREVALENT model is retrained on indoor image-text pairs that we have prepared, as shown in Figure \ref{fig:fig2}. Firstly, we freeze all parameters of the PREVALENT backbone model, which could not be updated during training. Then we add additional visual prompts on the first layer, and an MLP head after the $N\mbox{-}th$ layer and visual prompts are learnable during training. We apply a cross-entropy loss to optimize only soft visual prompts and the linear head by prompt tuning. With such a low-consumption auxiliary classification task, the visual prompts are expected to inject the knowledge of object-level and scene-level indoor image semantics into the PREVALENT model. The text encoder and the updated visual encoder process the instruction and the current observation information respectively, and the across-modal module predicts the next action.

\section{Experiment}
\label{sec:typestyle}

We show the experiment setup in Sec. \ref{sec:setup} and compare DAP with existing SOTA methods in Sec. \ref{sec:sota}. In additoin, we further verify that DAP is model-agnostic in Sec. \ref{ssec:3.3}.

\subsection{Experimental Setup}
\label{sec:setup}
Our training process includes updating the pretrained model and adapting it to downstream VLN tasks. Without loss of generality, our baseline agent follows the architecture of RecBERT~\cite{DBLP:conf/cvpr/Hong0QOG21}, which initializes from the pretrained model OSCAR~\cite{li2020oscar} learned from out-of domain datasets or PREVALENT \cite{hao2020towards} learned from VLN datasets. We retrain the PREVALENT model with about 1000 in-domain image-text pairs generated by us, and perform prompt tuning for 20 epochs with batch size 10. The number of soft visual prompts is 10. After that, we adapt DAP to the downstream generative VLN task with fine-tuning. For R2R, we train the agent based on RecBERT~\cite{DBLP:conf/cvpr/Hong0QOG21} initialized from PREVALENT for 300,000 iterations and the batch size is 8. For REVERIE, we train the agent for 200,000 iterations with batch size 8. All experiments are conducted on a single NVIDIA 3090 GPU.

\subsection{Comparison to State-of-the-Art Methods}
\label{sec:sota}

Table \ref{tab:table1} compare the performance of classical methods on the R2R. Compared to the baseline model RecBERT \cite{DBLP:conf/cvpr/Hong0QOG21}, DAP achieves 65\% SR \textbf{(+2\%)} and 59\% SPL \textbf{(+2\%)} on the validation unseen. On the test unseen split, we achieve 64\% SR \textbf{(+1\%)} and 59\% SPL \textbf{(+2\%)}. The large improvement suggests that domain-aware knowledge for pretrained models benefits the learning of the VLN agent. DAP has the best results across the main metrics, even compared against some newest words such as ADAPT\cite{DBLP:conf/cvpr/Lin0CLLL22} and GRVLN-BERT\cite{DBLP:conf/swarm/ZhangQZZYWWX23}.

We compare DAP with these methods on the REVERIE dataset, as shown in Table \ref{tab:table2}. Compared to RecBERT (init. PREVALNRT) \cite{DBLP:conf/cvpr/Hong0QOG21}, we achieve 1.67\% improvement on RGS and 2.05\% improvement on RGSPL on the validation unseen split. On the test unseen split, we achieve 1.03\% improvement on RGS and 1.27\% improvement on RGSPL. This suggests that DAP is better for locating target objects.

\definecolor{Gray}{gray}{0.94}
\begin{table*}[ht]
\begin{center}
\vspace{-2mm}
\caption{Comparison with the SOTA methods on R2R dataset.}
\resizebox{0.80\textwidth}{!}{
\begin{tabular}{lrrrrrr>{\columncolor{Gray}}r>{\columncolor{Gray}}rrr>{\columncolor{Gray}}r>{\columncolor{Gray}}r}
    \toprule 
    \multicolumn{1}{c}{\multirow{2}{*}{Agent}} & \multicolumn{4}{c}{Val Seen} & \multicolumn{4}{c}{Val Unseen} & \multicolumn{4}{c}{Test Unseen} \\
    \cmidrule(r){2-5} \cmidrule(r){6-9} \cmidrule(r){10-13}
    \multicolumn{1}{c}{} & \multicolumn{1}{c}{TL} & \multicolumn{1}{c}{NE$\downarrow$} & \multicolumn{1}{c}{SR$\uparrow$} & \multicolumn{1}{c}{SPL$\uparrow$} & \multicolumn{1}{c}{TL} & \multicolumn{1}{c}{NE$\downarrow$} & \multicolumn{1}{c}{SR$\uparrow$} & \multicolumn{1}{c}{SPL$\uparrow$} & \multicolumn{1}{c}{TL} & \multicolumn{1}{c}{NE$\downarrow$} & \multicolumn{1}{c}{SR$\uparrow$} & \multicolumn{1}{c}{SPL$\uparrow$} \\
    \midrule
    Random                & 9.58  & 9.45 & 16 & -    & 9.77  & 9.23 & 16 & -    & 9.89  & 9.79 & 13 & 12 \\
    Human                 & -     & -    & -    & -    & -     & -    & -    & -    & 11.85 & 1.61 & 86 & 76 \\
    \midrule

    
    PRESS \cite{DBLP:conf/emnlp/LiLXBCGSC19}~
    & 10.57 & 4.39 & 58 & 55 & 10.36 & 5.28 & 49 & 45 & 10.77 & 5.49 & 49 & 45 \\
    EnvDrop \cite{DBLP:conf/naacl/TanYB19}~
    & 11.00 & 3.99 & 62 & 59 & 10.70 & 5.22 & 52 & 48 & 11.66 & 5.23 & 51 & 47 \\
    PREVALENT \cite{hao2020towards}~
    & 10.32 & 3.67 & 69 & 65 & 10.19 & 4.71 & 58 & 53 & 10.51 & 5.30 & 54 & 51 \\
    EnvDrop+REM \cite{DBLP:conf/iccv/0002ZCLGS21}~
    & 11.13 & 3.14 & 70 & 66 & 14.84 & 4.99 & 53 & 48 & 10.73 & 5.40 & 54 & 50 \\
     AuxRN \cite{zhu2020vision}~
    & -     & 3.33 & {70} & \ 67 & -     & 5.28 & 55 & 50 & -     & 5.15 & 55 & 51 \\
    ORIST \cite{DBLP:conf/iccv/QiPH0H021}~
    & - & - & - & - & 10.90 & 4.72 & 57 & 51 & 11.31 & 5.10 & 57 & 52 \\
    NvEM \cite{DBLP:conf/mm/AnQHWWT21}~  & 11.09 & 3.44 & 69 & 65 & 11.83 & {4.27} & {60} & {55} & 12.98 & {4.37} & {58} & {54} \\
    EnvDrop+SEvol \cite{DBLP:conf/cvpr/ChenGMZ022} & 12.55     &  3.70   &  61   &  57    &  14.67   &  4.39   &  59 &  53   &  14.30    &  \high{\textbf{3.70}}  &   59  &  55   \\  
    NvEM+SEvol \cite{DBLP:conf/cvpr/ChenGMZ022}
    & 11.97     &  3.56   &  67   &  63    &  12.26   &  \ 3.99   &  \ 62 &  57   &  13.40    &  \ 4.13  &   \ 62  &  \ 57   \\

    ProbES \cite{DBLP:conf/acl/LiangZLXL22}
    & 10.75     &  2.95   &  73   &  69    &  11.58   &  \ 4.03   &  \ 61 &  \ 55   &  12.43    &  \ 4.20  &   \ 62  &  \ 56   \\
     ADAPT \cite{DBLP:conf/cvpr/Lin0CLLL22}
    & 11.39     & 2.70   &  74   &  69    &  12.33   &   3.66   &   66 &   59   &  13.16    &  4.11 &   \ 63  &  \ 57   \\
    GRVLN-BERT \cite{DBLP:conf/swarm/ZhangQZZYWWX23}
    & 11.08     & 2.58   &  75   &  71    &  12.49   &   3.81   &   62 &   56   &  12.78    &  3.96 &   \ 63  &  \ 57   \\
    \hline
    RecBERT (init. OSCAR)~\cite{DBLP:conf/cvpr/Hong0QOG21}	& 10.79 & 3.11 & 71 & 67 & 11.86 & 4.29 & 59 & 53 & 12.34 & 4.59 & 57 & 53 \\
    RecBERT (init. PREVALENT) \cite{DBLP:conf/cvpr/Hong0QOG21} ~ &  11.13 & 2.90 & 72 & 68 & 12.01 & 3.93 & 63 & 57 & 12.35 & 4.09 & 63 & 57 \\
   
    \midrule
    DAP(Ours)~ & \ 10.98 & \high{\textbf{2.53}} & \high{\textbf{76}} & \high{\textbf{72}} &  12.12 & \high{\textbf{3.62}} & \high{\textbf{65}} & \high{\textbf{59}} & 12.07 & {3.95} & \high{\textbf{64}} & \high{\textbf{59}} \\

    \bottomrule
\end{tabular}
}
\vspace{-2mm}
\label{tab:table1}
\end{center}
\end{table*}

\begin{table*}[ht]
\caption{Comparison of the agent performance of navigation and remote referring expression on REVERIE.}
\begin{center}
  \resizebox{\linewidth}{!}{
  \begin{tabular}{l|rrrrrr|rrrrrr|rrrrrr}
  \hline
  \hline
  \multicolumn{1}{c}{\multirow{3}{*}{Methods}}  & \multicolumn{6}{|c|}{REVERIE Validation Seen} &\multicolumn{6}{c|}{REVERIE Validation Unseen} & \multicolumn{6}{c}{REVERIE Test Unseen} \Tstrut\\
  \cline{2-19}& \multicolumn{4}{c|}{Navigation}  & \multicolumn{1}{c}{\multirow{2}{*}{RGS$\uparrow$}}&\multicolumn{1}{c|}{\multirow{2}{*}{RGSPL$\uparrow$}} & \multicolumn{4}{c|}{Navigation}  & \multicolumn{1}{c}{\multirow{2}{*}{RGS$\uparrow$}}&\multicolumn{1}{c|}{\multirow{2}{*}{RGSPL$\uparrow$}} & \multicolumn{4}{c|}{Navigation}   & \multicolumn{1}{c}{\multirow{2}{*}{RGS$\uparrow$}}&\multicolumn{1}{c}{\multirow{2}{*}{RGSPL$\uparrow$}} \\
  \cline{2-5} \cline{8-11} \cline{14-17}  & \multicolumn{1}{c}{SR$\uparrow$} & \multicolumn{1}{c}{OSR$\uparrow$} & \multicolumn{1}{c}{SPL$\uparrow$}  & \multicolumn{1}{c|}{TL} &  &  & \multicolumn{1}{c}{SR$\uparrow$} & \multicolumn{1}{c}{OSR$\uparrow$} & \multicolumn{1}{c}{SPL$\uparrow$} &\multicolumn{1}{c|}{TL} & & & \multicolumn{1}{c}{SR$\uparrow$} & \multicolumn{1}{c}{OSR$\uparrow$} & \multicolumn{1}{c}{SPL$\uparrow$} & \multicolumn{1}{c|}{TL} &  & \Tstrut\\
  \hline 
  Random &2.74 &8.92& 1.91 & \multicolumn{1}{c|}{11.99} & 1.97 & 1.31 &1.76& 11.93 &1.01& \multicolumn{1}{c|}{10.76}  & 0.96 & 0.56 &  2.30  &8.88 & 1.44&  \multicolumn{1}{c|}{10.34} &1.18 & 0.78 \\
  Human & -- & --  & -- & \multicolumn{1}{r|}{--} & -- & -- & -- & --  & --  & \multicolumn{1}{r|}{--} & -- & -- & 81.51 & 86.83  & 53.66 & \multicolumn{1}{r|}{21.18} & 77.84 &  51.44 \\
  \hline
  
  RCM \cite{wang2019reinforced} & 23.33 & 29.44 & 21.82& \multicolumn{1}{r|}{10.70} & 16.23 & 15.36 & 9.29 & 14.23 & 6.97 &\multicolumn{1}{r|}{11.98} & 4.89 & 3.89 & 7.84 & 11.68 & 6.67 & \multicolumn{1}{r|}{10.60} & 3.67 & 3.14\\
  SMNA \cite{DBLP:conf/iclr/MaLWAKSX19} & 41.25& 43.29  & 39.61& \multicolumn{1}{r|}{7.54}   & 30.07 &  28.98 & 8.15 & 11.28 &6.44 & \multicolumn{1}{r|}{9.07} & 4.54& 3.61 & 5.80& 8.39 & 4.53 &\multicolumn{1}{r|}{9.23}   & 3.10& 2.39 \\
  FAST-Short \cite{ke2019tactical} & 45.12& 49.68 &40.18& \multicolumn{1}{r|}{13.22}  &31.41 & 28.11 & 10.08 & 20.48 & 6.17 & \multicolumn{1}{r|}{29.70}  & 6.24 & 3.97 & 14.18 & 23.36 & 8.74 & \multicolumn{1}{r|}{30.69}  & 7.07 & 4.52 \\
  FAST-MATTN \cite{qi2020reverie} & \high{50.53} & \high{55.17} & \high{45.50} & \multicolumn{1}{r|}{{16.35}} & \high{31.97} & \high{29.66} & \ 14.40 & \high{28.20} &  7.19 & \multicolumn{1}{r|}{{45.28}}  & \ 7.84 & \ 4.67 & \ 19.88 & \ 30.63 & \ 11.61 & \multicolumn{1}{r|}{{39.05}} & \ 11.28 & \ 6.08 \\
  ProbES~\cite{DBLP:conf/acl/LiangZLXL22} & \ 46.52 & \ 48.49 & \ 42.44 & \multicolumn{1}{r|}{{13.59}} & \ 33.66 & \ 30.86 & \high{27.63} & \ 33.23 & \high{22.75} & \multicolumn{1}{r|}{{18.00}} & \high{16.84} & \high{13.94} & \high{24.97} & \ 28.23 & \high{20.12} & \multicolumn{1}{r|}{{17.43}} & \high{15.11} & \high{12.32} \\
  \hline
  RecBERT (init. OSCAR) \cite{DBLP:conf/cvpr/Hong0QOG21}& \ 39.85 & \ 41.32 & \ 35.86 & \multicolumn{1}{r|}{{12.85}} & \ 24.46 & \ 22.28 & \high{25.53} & \ 27.66 & \high{21.06} & \multicolumn{1}{r|}{{14.35}} & \high{14.20} & \high{12.00} & \high{24.62} & \ 26.67 & \high{19.48} & \multicolumn{1}{r|}{{14.88}} & \high{12.65} & \high{10.00} \\
  RecBERT (init. PREVALENT) \cite{DBLP:conf/cvpr/Hong0QOG21} & \ 51.79 & \ 53.90 & \ 47.96 & \multicolumn{1}{r|}{{13.44}} & \ 38.23 & \ 35.61 & \high{30.07} & \ 35.02 & \high{24.90} & \multicolumn{1}{r|}{{16.78}} & \high{18.77} & \high{15.27} & \high{29.61} & \ 32.91 & \high{23.99} & \multicolumn{1}{r|}{{15.86}} & \high{16.05} & \high{13.51} \\
  \hline
  DAP (Ours) & \textbf{53.81} & \textbf{55.60} & \textbf{50.57} & \multicolumn{1}{r|}{{13.58}} & \textbf{40.38} & \textbf{37.62} & \high{\textbf{32.17}} & \textbf{37.21} & \high{\textbf{27.30}} & \multicolumn{1}{r|}{16.32} & \high{\textbf{20.44}} & \high{\textbf{17.32}} & \high{\textbf{30.26}} & \textbf{33.15} & \high{\textbf{24.07}} & \multicolumn{1}{r|}{15.37} & \high{\textbf{17.08}} & \high{\textbf{14.78}} \\
  \hline \hline
\end{tabular}}
\end{center}

\label{tab:table2}
\end{table*}

\begin{table}
    \caption{The results of using soft visual prompts on different pretrained models on R2R.}

	\fontsize{12}{12}\selectfont
		
			\resizebox{1.0\linewidth}{!}{
	{\renewcommand{\arraystretch}{1.5}
 \begin{tabular}{c||c|c|c|c|c|c}
			
			 \specialrule{.1em}{.05em}{.05em}
		
			\multirow{2}{*}{Methods}&\multicolumn{3}{c|}{Val seen}&\multicolumn{3}{c}{Val Unseen}\cr\cline{2-7}
			&NE  $\downarrow$ &SR $\uparrow$ &SPL $\uparrow$ &NE $\downarrow$ &SR $\uparrow$ &SPL $\uparrow$\cr
			\hline
            \hline
   
            RecBERT (init.  OSCAR) \cite{DBLP:conf/cvpr/Hong0QOG21}& 3.11&71.11&67.23&4.29&58.71&53.41\\
            \hline
             DAP (init. OSCAR)&\textbf{2.86}&\textbf{73.21}&\textbf{68.14}& \textbf{4.11}&\textbf{62.13}&\textbf{56.59}\\ 
        
           \hline
           \hline
           RecBERT (init. PREVALENT) \cite{DBLP:conf/cvpr/Hong0QOG21}&2.90&72.18&67.72&3.93&62.75&56.84\\
         \hline
         DAP (init. PREVALENT)&\textbf{2.53}&\textbf{75.58}&\textbf{71.52}&\textbf{3.62}&\textbf{64.51}&\textbf{58.63}\\
         

      	 \specialrule{.1em}{.05em}{.05em}
    \end{tabular}}}

	\label{tab:table3}
\end{table}

\subsection{Effects of Visual Prompts Based on Different Pretrained Models.}
\label{ssec:3.3}

As shown in Table \ref{tab:table3}, we further investigate whether the in-domain pretraining works for different visual-and-language pretrained models by additionally conducting the experiment on another pretrained model named OSCAR \cite{li2020oscar}. On the validation unseen split, compared with the RecBERT (init. OSCAR) \cite{DBLP:conf/cvpr/Hong0QOG21}, DAP boosts SR and SPL from 58.71\% and 53.41\% to 62.13\% \textbf{(+3.42\%)} and 56.59\% \textbf{(+3.18\%)} respectively. By comparing with the RecBERT (init. PREVALENT) \cite{DBLP:conf/cvpr/Hong0QOG21}, DAP improves SR and SPL from 62.75\% and 56.84\% to 64.51\% \textbf{(+1.76\%)} and 58.63\% \textbf{(+1.79\%)} respectively. It can be noted that the soft visual prompts yield a larger improvement on OSCAR than PREVALENT. The reason is that OSCAR is trained on datasets crawled from the web and leaves more headroom for enhancement by in-domain prompt tuning, whereas PREVALENT is trained on VLN datasets. The results show that soft visual prompts play an important role in reducing the gap between pretraining datasets and VLN datasets. This experiment further verifies that soft visual prompts are independent of specific pre-trained models, i.e., DAP is model-agnostic.

\section{Conclusion}
\label{sec:print}
In this work, we propose a novel and model-agnostic framework named Domain-aware Prompt learning (\emph{DAP}), which prompts the VLN agent with the capability of recognizing objects and scenes in visual perceptions. DAP achieves better results compared to previous SOTA methods on both R2R and REVERIE, and the experiment shows the effect of soft visual prompts in different pretrained models. We believe that DAP also can benefit future studies in other vision-and-language tasks. Given that the proposed approach is focused on pretraining a representation model, it is worthwhile to investigate the effect of combining it with others on navigation inference.

\subsection*{Acknowledgement}

This research was supported partially by the National Natural Science Fund of China (Grant Nos. 62306329 and 62103425, and the Natural Science Fund of Hunan Province (Grant Nos. 2023JJ40676 and 2022JJ40559).


\clearpage

\bibliographystyle{IEEEbib}
\bibliography{strings,refs}

\end{document}